%% file: main.tex
\documentclass{article} 
\usepackage{iclr2025_delta,times}
\usepackage{graphicx} 
\usepackage{amsmath}
\usepackage{hyperref}
\usepackage{amssymb}
\usepackage{xcolor}
\usepackage{float}
\usepackage{caption}
\usepackage{algorithm}
\usepackage{algorithmic}
\usepackage{epstopdf}
\usepackage{glossaries}
\usepackage{cleveref}

\input{math_commands.tex}

\input{math}
\input{acronyms}

\title{INFO-SEDD: Continuous Time Markov Chains as Scalable Information Metrics Estimators}

\author{Alberto Foresti, Giulio Franzese, Pietro Michiardi \\ EURECOM\\
\texttt{\{alberto.foresti,giulio.franzese,pietro.michiardi\}@eurecom.fr}
}

\iclrfinalcopy

\begin{document}

\maketitle

\begin{abstract}
\input{abstract}

\end{abstract}

\section{Introduction}
\input{introduction}

\section{Methodology}\label{sec:method}
\input{klest}

\section{\infosedd}\label{sec:infosed}
\input{infosedd}

\section{Experimental Validation}
\input{experiments}

\section{Conclusion}
\input{conclusions}

\bibliography{main}
\bibliographystyle{iclr2025_delta}

\appendix
\input{appendix}

\end{document}

%% file: math_commands.tex
\usepackage{amsmath,amsfonts,bm}

\def\eqref#1{equation~\ref{#1}}

\def\1{\bm{1}}

\def\rvx{{\mathbf{x}}}
\def\rvy{{\mathbf{y}}}

\DeclareMathAlphabet{\mathsfit}{\encodingdefault}{\sfdefault}{m}{sl}
\SetMathAlphabet{\mathsfit}{bold}{\encodingdefault}{\sfdefault}{bx}{n}

\def\sN{{\mathbb{N}}}

\newcommand{\KL}{D_{\mathrm{KL}}}



%% file: math.tex
\newcommand{\g}{\,|\,}

\newcommand{\prob}{\text{Pr}}

\DeclareRobustCommand{\KL}[2]{\ensuremath{\textsc{kl}\left[#1\;\|\;#2\right]}}

\newcommand{\bbR}{\mathbb{R}}

\newcommand{\defeq}{\stackrel{\text{\tiny def}}{=}}

\newcommand{\pjoint}{{p}}
\newcommand{\pmarginal}{m}

\newcommand{\fpjoint}{\overset{\rightarrow}{\pjoint}}

\newcommand{\fpmarginal}{\overset{\rightarrow}{\pmarginal}}

\newcommand{\fp}{\overset{\rightarrow}{p}}
\newcommand{\bp}{\overset{\leftarrow}{p}}

\newcommand{\fq}{\overset{\rightarrow}{q}}
\newcommand{\bq}{\overset{\leftarrow}{q}}

\newcommand{\fu}{\overset{\rightarrow}{u}}

\newcommand{\fratesp}{\overset{\rightarrow}{Q}}
\newcommand{\bratesp}{\overset{\leftarrow}{Q}}

\newcommand{\fratestokp}{{\overset{\rightarrow}{Q}}^{\text{tok}}}

\newcommand{\support}{\chi}

\newcommand{\fproc}{\overset{\rightarrow}{X}}
\newcommand{\bproc}{\overset{\leftarrow}{X}}

\newcommand{\fprocy}{\overset{\rightarrow}{Y}}

\newcommand{\xbar}{\bar{x}}

\newcommand{\pjump}{\prob(\fprocy_s\text{ jumps to }\absorb\text{ in }[0,t])}

\newcommand{\ones}{\mathbf{1}}

\newcommand{\expected}{\mathbb{E}}

\newcommand{\boperator}{\mathcal{B}}

\newcommand{\scorep}{s^p_\theta}
\newcommand{\scoreq}{s^q_\theta}

\newcommand{\scorefn}{s_\theta}

\newcommand{\hamming}{d_{\text{hamming}}}

\DeclareMathAlphabet{\pazocal}{OMS}{zplm}{m}{n}

\newcommand{\cumnoise}{\overline{\sigma}}
\newcommand{\absorb}{\emptyset}
\newcommand{\ratio}{\frac{1}{N(e^{\cumnoise(t)}-1)}}

\newcommand{\boltzmann}{k_b}
\newcommand{\pf}{\lambda}

\newcommand{\infosedd}{\textsc{info-sedd}}

%% file: acronyms.tex
\newacronym{ODE}{\textsc{ode}}{Ordinary Differential Equation}
\newacronym{CTMC}{\textsc{ctmc}}{Continuous Time Markov Chain}
\newacronym{KL}{\textsc{kl}}{Kullback-Leibler}
\newacronym{MLP}{\textsc{MLP}}{Multi-Layer-Perceptron}

%% file: abstract.tex
Information-theoretic quantities play a crucial role in understanding non-linear relationships between random variables and are widely used across scientific disciplines. However, estimating these quantities remains an open problem, particularly in the case of high-dimensional discrete distributions. Current approaches typically rely on embedding discrete data into a continuous space and applying neural estimators originally designed for continuous distributions, a process that may not fully capture the discrete nature of the underlying data.

We consider Continuous-Time Markov Chains (\textsc{CTMCs}), stochastic processes on discrete state-spaces which have gained popularity due to their generative modeling applications. In this work, we introduce \infosedd, a novel method for estimating information-theoretic quantities of discrete data, including mutual information and entropy. Our approach requires the training of a single parametric model, offering significant computational and memory advantages. Additionally, it seamlessly integrates with pretrained networks, allowing for efficient reuse of pretrained generative models.

To evaluate our approach, we construct a challenging synthetic benchmark. Our experiments demonstrate that \infosedd\ is robust and outperforms neural competitors that rely on embedding techniques. Moreover, we validate our method on a real-world task: estimating the entropy of an Ising model. Overall, \infosedd\ outperforms competing methods and shows scalability to high-dimensional scenarios, paving the way for new applications where estimating MI between discrete distribution is the focus.

%% file: introduction.tex
Information theoretic quantities represent a powerful tool to understand non-linear relationships between random variables \citep{shannon1948mathematical, mackay2003information} and find wide range of applications in scientific fields \citep{Karbowski_2024, Eckford_2016}. Mutual information, in particular, has become an established metric in machine learning \citep{bell1995information, stratos2018mutual, belghazi2018mine, oord2019representation, hjelm2019learning}, both for training models \citep{alemi2019deep, chen2016infogan, zhao2018information} and at inference time \citep{alemi2019gilbo, huang2020evaluating}.

Estimating information theoretic quantities remains an open problem, and different paradigms for their estimation emerged. Classical parametric and non-parametric methods \citep{pizer1987adaptive, moon1995estimation, kraskov2004estimating, gao2015efficient} have been recently superseded by variational approaches \citep{barber2004algorithm, nguyen2007neurips, nowozin2016neurips, poole2019variational, wunder2021reverse, letizia2023variational, federici2023effectiveness} and neural estimators \citep{papamakarios2017masked, belghazi2018mine, oord2019representation, song2019, rhodes2020telescoping, letizia2022copula, brekelmans2023improving,franzese2023minde,butakov2024mutual}. Despite its practical importance, few estimators for high-dimensional \textbf{discrete} distributions have been proposed in the literature. While classical estimators for discrete random variables exist \citep{pinchas2024comparative}, their accuracy rapidly decrease with increasing dimensionality of the considered problem. Applications that would benefit from scalable estimators of mutual information quantities include, among others, DNA or peptide sequencing \citep{newcomb2021use, xiaadanovo}, text summarization \citep{darrin2024textttcosmicmutualinformationtaskagnostic} and neuroscience \citep{chai2009exploring}. Consequently, the development of new estimation techniques is of paramount importance for the broader scientific community.

The current approach to solve the problem of lack of a viable estimator for discrete, high dimensional scenarios, is to embed in a real space the discrete quantities and then adopt neural estimators conceived for continuous distributions. One recent example is \citep{lee2024benchmarksuiteevaluatingneural}, where it is showed that the embeddings of pretrained language models can provide meaningful representations to estimate information theoretic quantities in unstructured data. However such a process may not fully capture the discrete nature of the underlying data and might suffer from several limitations, such as the necessity to consider embeddings which are application specific.

One extremely promising class of estimators which has been recently been considered in the continuous state space settings has its roots in generative diffusion models \cite{song2021a,ho2020}. While these generative models have been successfully considered in continuous settings for estimating information metrics \citep{franzese2023minde,kong2023information,bounouas}, a discrete state space adaptation is currently not available. In this work we fill this important gap and present \infosedd, a novel method for estimating information theoretic quantities of discrete data using \glspl{CTMC} \citep{lou2024discrete}. These stochastic processes have recently saw a surge in popularity due to the associated generative modelling applications as direct counterpart of the continuous state space models \cite{song2021a,ho2020}. Their fundamental working principle is the reversal of a perturbation process which starts from clean data from a given distribution and converge to uninformative noise. The workhorse of these approaches is the \textit{score function}, which contains information about the probability distributions associated to the \glspl{CTMC} at different time instants. Our proposed method, \infosedd, builds upon the same fundamental mathematical framework, extending it via Dynkin's lemma \citep{hanson2007applied}, and leverages score functions to compute key information-theoretic metrics, such as mutual information between two random variables and the entropy of a given distribution. By carefully selecting perturbation processes, our approach requires training only a single parametric model to compute mutual information across arbitrary subsets of variables. Furthermore, \infosedd\ seamlessly integrates with pretrained networks, enabling the reuse of computational resources already expended in training generative models. 

To rigorously evaluate our method, we design a benchmark that presents challenges across three critical dimensions: (1) the support size of the random variables, which defines the range of possible values each variable can take; (2) the dimensionality of the variable representations, referring to the number of components each variable contains; and (3) the mutual information value, which captures the complexity of dependencies between variables. Our results demonstrate that \infosedd\ is both robust and consistently outperforms neural estimation methods that rely on embedding techniques.
Beyond synthetic benchmarks, we further assess the practical utility of our approach by applying it to the real-world problem of estimating the entropy of the Ising model \citep{onsager1944crystal}. The Ising model is a paradigmatic example of a complex system with broad applications in statistical physics, neuroscience, and machine learning. Crucially, it provides a well-characterized ground truth for entropy, making it an ideal test case for evaluating the accuracy of information-theoretic estimators in high-dimensional discrete distributions.
Our method achieves precise entropy estimates in this challenging setting, reinforcing \infosedd\ as a promising and reliable estimator for complex discrete data distributions.

%% file: klest.tex
In this Section, we explore the relationship between \glspl{CTMC} over discrete state spaces \citep{anderson2012continuous} and the computation of \gls{KL} divergences. First, in \Cref{subsec:ctmc}, we provide a brief introduction to the fundamentals of \glspl{CTMC}, emphasizing their time-reversal properties and parametric approximations \citep{lou2024discrete}. Then, in \Cref{subsec:klfor}, we demonstrate how these processes can be adapted for divergence estimation, specifically by analyzing two processes that share the same generator but differ in their initial conditions.

\subsection{Preliminaries}\label{subsec:ctmc}
Consider a \gls{CTMC} $\fproc_t, t \in[0,T]$, defined over a finite state space $\support = \left\{1,\hdots,N\right\}$ and  specified by the infinitesimal generators $\fratesp_t: [0,T]\rightarrow \mathbb{R}^{N\times N}$, where the diagonal entries statisfy $\fratesp_t(x,x)=-\sum_{x\neq y}\fratesp_t(x,y)$, with $\fratesp_t(x,y)\geq 0, x\neq y$. As established in \citep{anderson2012continuous}, the time evolution of the probability distribution $Pr(X_t=i)\defeq \fp_t(i)$, satisfies the following \gls{ODE} 
\begin{equation}\label{for_ode}
\fp_t =\fp_0+\int_0^t \fp_s\fratesp_s ds,
\end{equation}
where the initial conditions of the process $\fp_0$ determine the distribution $\fp_t$ at any time $t$. 

A key property of \glspl{CTMC} is that their time-reversed counterpart, recently utilized in generative modeling \citep{lou2024discrete}, also follows a \gls{CTMC} but with a different set of transition matrices. More precisely, defining the time-reversed process as $\bp_t\defeq\fp_{T-t}$, it evolves according to \citep{lou2024discrete,sun2023scorebasedcontinuoustimediscretediffusion}:
\begin{align}\label{eq:bctmc}
    \bp_t =\fp_T+\int_0^t \bratesp_s\bp_s ds,
\end{align}
where the reverse-time transition matrices $\bratesp_t$ relate to the forward transition matrices as follows:

\begin{align}\label{eq:ratesdef}
    \bratesp_t(y,x) &= \left(\frac{\fp_{T-t}(y)}{\fp_{T-t}(x)} \fratesp_{T-t}(x,y)\right) (1-\delta(x,y))+\left(-\sum_{y\neq x}\bratesp_t(y,x)\right)\delta(x,x)
\end{align}

Under appropriate technical conditions on $\fratesp_t$ \citep{lou2024discrete} the terminal distribution $\fp_T$ converges to a known reference distribution $\pi$, which is independent of the initial distribution $\fp_0$. This property enables sampling from $\fp_0$ by simulating a \gls{CTMC} with appropriately chosen generators \citep{sun2023scorebasedcontinuoustimediscretediffusion, kelly1981reversibility}. However, other than simple and uninteresting scenarios, exact knowledge of the quantities $\frac{\fp_{T-t}(y)}{\fp_{T-t}(x)}$ is out of reach. A practical solution is to substitute in this numerical integration a parametric function $\scorep(x,t)_y$, whose parameters are optimized according to \citet{lou2024discrete}:

{\small\begin{equation}\label{eq:lossfn}
    \mathcal{L}(\theta) = \expected\left[\int_0^T\sum_{y\neq \fproc_t}\fratesp_t(\fproc_t,y)\left(\scorep(\fproc_t,t)_y-\frac{\fp_t(y|\fproc_0)}{\fp_t(\fproc_t|\fproc_0)}\log\scorep(\fproc_t,t)_y+K\left(\frac{\fp_t(y|\fproc_0)}{\fp_t(\fproc_t|\fproc_0)}\right)\right)dt\right]
\end{equation}}

Where $\fp_t(\cdot|x_0)$ is a known perturbation kernel, obtained from \Cref{for_ode} with the deterministic initial distribution centered in $x_0$, and $K(a)=a(\log a-1)$. Whenever the context is clear, we simplify the notation for the parametric score in the remainder of the paper and denote 
$\scorep(\fproc_t)_y$ instead of $\scorep(\fproc_t,t)_y$.

\subsection{\gls{KL} Divergences via \glspl{CTMC}}\label{subsec:klfor}

In this work, we leverage the \gls{CTMC} framework to compute the KL divergence between two probability distributions $\fp_0$ and $\fq_0$ defined over the same support $\support$, expressed as $\KL{\fp_0}{\fq_0}$. To achieve this, we construct two Markov chains that differ only in their initial conditions: one initialized from $\fp_0$ whose time evolution follows \Cref{eq:bctmc}, and the other initialized from $\fq_0$, which evolves analogously by substituting $\fq_t$ for $\fp_t$ in \Cref{eq:bctmc}. Since the KL divergence satisfies $\KL{\fp_0}{\fq_0}=\expected\left[\log\frac{\fp_0}{\fq_0}(\fproc_0)\right]$, we can equivalently express it as
\begin{equation}\label{eq:kl_expressions}
    \expected\left[\log\frac{\fp_0}{\fq_0}(\bproc_T)\right] =\expected\left[\log\frac{\bp_T}{\bq_T}(\bproc_T)\right] = \expected \left[\expected\left[\log\frac{\bp_T}{\bq_T}(\bproc_T)\bigg|\bproc_0\right]\right].
\end{equation}

The last term in \eqref{eq:kl_expressions} can be rewritten using Dynkin's formula \cite{hanson2007applied}, which states that for a generic function \( f:\support\times[0,T]\to\bbR \), we have:

\begin{equation}\label{eq:dynkin}
    \expected\left[f(\bproc_T,T)\bigg|\bproc_0\right]- f(\bproc_0,0)=\expected\left[\int_0^T\frac{\partial f}{\partial t}(\bproc_t,t)+\boperator[f](\bproc_t,t)dt\bigg|\bproc_0\right]
\end{equation}

where \( \boperator \) is the \textit{backward operator}, defined in our setting as:

\begin{equation}
    \boperator[f](x,t) = \sum_{y\neq x}\bratesp_t(y,x)(f(y)-f(x)).
\end{equation}

By combining the result from \Cref{eq:dynkin} with \Cref{eq:kl_expressions}, we obtain the following expression for the KL divergence between discrete distributions \( \KL{\fp_0}{\fq_0} \):

\begin{align}\label{eq:infosedd}
&\expected\left[\int_0^T\sum_{y\neq \fproc_t} \fratesp_{t}(\fproc_t,y)\left(K\left(\frac{\fp_t(y)}{\fp_t(\fproc_t)}\right)+\frac{\fq_t(y)}{\fq_t(\fproc_t)}-\frac{\fp_t(y)}{\fp_t(\fproc_t)}\log\frac{\fq_t(y)}{\fq_t(\fproc_t)}\right)dt\right]
\end{align}

where \( K(a)=a(\log(a)-1) \). The missing term \( \expected\left[\log{\frac{\bp_0}{\bq_0}}(\bproc_0)\right] \) is omitted, as both \( \bp_0 \) and \( \bq_0 \) converge to \( \pi \) \citep{lou2024discrete}.

While \Cref{eq:infosedd} provides a complete formulation for estimating the divergence of interest, in practical applications—similar to those in generative modeling \citep{lou2024discrete, ren2024discrete, holderrieth2024generator}—the key quantities \( \frac{\fp_{t}(y)}{\fp_{t}(x)} \) and \( \frac{\fq_{t}(y)}{\fq_{t}(x)} \) are not directly accessible. To address this, we adopt the approach of \citet{lou2024discrete}, replacing these unknown ratios with parametric approximations optimized via \Cref{eq:lossfn}. This leads to the construction of the estimator:

\begin{align}\label{eq:infosedd-param} 
&\int_0^T\expected\left[\sum_{y\neq \fproc_{t}} \fratesp_{t}(\fproc_{t},y)\left(K\left(\scorep(\fproc_{t})_y\right)+\scoreq(\fproc_{t})_y-\scorep(\fproc_{t})_y\log\scoreq(\fproc_{t})_y\right)\right] dt
\end{align}

where \( \scorep(\bproc_t)_y \approx \frac{\bp_t(y)}{\bp_t(\bproc_t)} \) and \( \scoreq(\bproc_t)_y\approx\frac{\bq_t(y)}{\bq_t(\bproc_t)} \) serve as parametric approximations of the true ratios. Estimating \Cref{eq:infosedd-param} using Monte Carlo techniques is conceptually straightforward: we sample time instants \( t \) uniformly in \( [0,T] \), simulate the forward process \( X_t \), and compute the required quantities using the parametric scores. However, as we will discuss in the next Section, the estimator in its general form is not scalable for computing information metrics, which is the primary focus of this work.

%% file: infosedd.tex
In this section, we introduce our mutual information estimator, \infosedd, which is based on \Cref{eq:infosedd-param}. Given two random variables, $\fproc_0$ and $\fprocy_0$, mutual information can be expressed in terms of the KL-divergence between the joint distribution $\fpjoint_0$ and the product of the marginals, defined as $\fpmarginal_0 = \fp^X_0 \otimes \fp^Y_0$, where $\otimes$ denotes the Kronecker product. This leads to the standard formulation: $I(\fproc_0,\fprocy_0) = \KL{\fpjoint_0}{\fpmarginal_0}$. However, this approach presents two key limitations, which we address in the following. 

First, in high dimensional applications, a naive implementation of \Cref{eq:infosedd-param} quickly becomes unfeasible. Indeed, the size of the number of entries of the matrix $\fratesp_t$ scales with $|\support|^2$, becoming quickly untractable. 
Fortunately, in many cases of interest, the random variables can be naturally decomposed into a structured sequence of $M$ subcomponents, each taking values from a discrete set of size  $N$ \citep{lou2024discrete, austin2021structured, campbell2022continuous}, i.e. $\fproc_0=[\fproc^1_0,\dots,\fproc^{M}_0]$, leading to a total state space of size $|\support|=N^M$.This structured decomposition enables the use of sparse rate matrices, which constrain the \gls{CTMC} to modify only one subcomponent at a time, significantly reducing computational complexity:\begin{equation}\label{eq:hamming_q}
\fratesp_t(x,y) = \delta(\hamming(x,y),1)\left(\sum_{i}(1-\delta(x^i,y^i))\fratestokp_t(x_i,y_i)\right),\quad x\neq y
\end{equation}
Since the summation in the inner expectation of \Cref{eq:infosedd-param} ( $\sum_{y\neq \fproc_{t}} \fratesp_{t}(\fproc_{t},y)\dots$ ) depends only on non-zero entries of the matrix $\fratesp_{t}(\fproc_{t},y)$, the formulation in \Cref{eq:hamming_q} greatly reduces the number of transitions that need to be considered.

Second, formulating mutual information as the KL divergence between the joint distribution and the product of marginals, $\KL{\fpjoint_0}{\fpmarginal_0}$, typically requires training two separate score models, each tailored to a specific distribution. However, a carefully chosen transition matrix can circumvent this requirement, allowing for a single unified model to be trained instead. In particular, selecting $\fratestokp_t=\sigma(t)\fratestokp_{\text{absorb}}$, with $\sigma(t)$ a fixed scalar function, and the absorbing matrix \footnote{This configuration adds an absorbing state, increasing the dimension of the support}   
\citep{lou2024discrete, campbell2022continuous, austin2021structured},
\begin{equation}
    \begin{array}{cc}
    \fratestokp_{\text{absorb}} = \begin{bmatrix}
    -1 & 0 & \cdots & 0 & 0 \\
    0 & -1 & \cdots & 0 & 0 \\
    \vdots & \vdots & \ddots & \vdots & \vdots \\
    0 & 0 & \cdots & -1 & 0 \\
    1 & 1 & \cdots & 1 & 0
    \end{bmatrix}
\end{array}
\end{equation}
ensures that the subcomponents can only transition into an absorbing state $\absorb$. 
This choice is crucial because it enables the computation of marginal scores using a model trained solely on the joint distribution. Specifically, as demonstrated in \Cref{sec:free_marginal},
\begin{equation}\label{eq:free_marginal}
    \frac{\fp_t(\fproc_t=x,\fprocy_t=\absorb)}{\fp_t(\fproc_t=x',\fprocy_t=\absorb)}=\frac{\fp^X_t(x)}{\fp^X_t(x')},\quad \frac{\fp_t(\fproc_t=\absorb,\fprocy_t=y)}{\fp_t(\fproc_t=\absorb,\fprocy_t=y')}=\frac{\fp^Y_t(y)}{\fp^Y_t(y')}.
\end{equation}

This result implies that a single score model trained on the joint distribution is sufficient for computing the marginal scores as well. By integrating these design choices, we now present the full formulation of \infosedd, whose pseudocode is detailed in \Cref{alg:infosedd}.

\begin{algorithm}[H]
\caption{\infosedd: Estimate $I(\rvx,\rvy)$}
\label{alg:infosedd}
\begin{algorithmic}[1]
{\small
    \REQUIRE Initial sample $[\fproc_0,\fprocy_0]\sim \fp_0$, score network $\scorefn$
    \STATE $t\sim u(0,T)$
    \COMMENT{Sample time uniformly}
    \STATE $[\fproc_t,\fprocy_t]\sim \fp_t(\cdot | [\fproc_0,\fprocy_0])$
    \COMMENT{Perturb data}
    \STATE $\hat{I}=0$
    \FOR{$i: \fproc^i_t=\absorb$}
    \FOR{$n\in [1:N]$}
    \STATE {\scriptsize $\tilde{X}=[\fproc^1_t,\dots,\fproc^{i-1}_t,n,\fproc^{i+1}_t,\dots,\fproc^{M}_t]$}
    \STATE {\scriptsize$\hat{I}+=T\sigma(t)\left(K\left(\scorefn([\fproc_t,\fprocy_t])_{[\tilde{X},\fprocy_t]}\right)+\scorefn([\fproc_t,\absorb])_{[\tilde{X},\absorb]}-\scorefn([\fproc_t,\fprocy_t])_{[\tilde{X},\fprocy_t]}\log\left(\scorefn([\fproc_t,\absorb])_{[\tilde{X},\absorb]}\right)\right)$}
    \ENDFOR
    \ENDFOR
    \FOR{$i: \fprocy^i_t=\absorb$}
    \FOR{$n\in [1:N]$}
    \STATE {\scriptsize$\tilde{Y}=[\fprocy^1_t,\dots,\fprocy^{i-1}_t,n,\fprocy^{i+1}_t,\dots,\fprocy^{M}_t]$}
    \STATE {\scriptsize$\hat{I}+=T\sigma(t)\left(K\left(\scorefn([\fproc_t,\fprocy_t])_{[\fproc_t,\tilde{Y}]}\right)+\scorefn([\absorb,\fprocy_t])_{[\absorb,\tilde{Y}]}-\scorefn([\fproc_t,\fprocy_t])_{[\fproc_t,\tilde{Y}]}\log\left(\scorefn([\absorb,\fprocy_t])_{[\absorb,\tilde{Y}]}\right)\right)$}
    \ENDFOR
    \ENDFOR
    \RETURN $\hat{I}$}
    \end{algorithmic}
\end{algorithm}

\subsection{Estimating entropy}
Notably, the proposed class of estimators can be readily adapted for entropy estimation. The entropy of a given distribution can be expressed in terms of the \gls{KL} divergence from the uniform distribution 
$\fu_0$, as follows: $H(\fp_0) = \log N -\KL{\fp_0}{\fu_0}$. Since the ratio $\frac{\fu_t(x)}{\fu_t(\absorb)}=\ratio$, with $\cumnoise(t)=\int_0^t\sigma(s)ds$ (see \cref{sec:absorb_ratio}), we can extend the formulation of \Cref{eq:infosedd-param} to derive \textsc{info-sedd-h}, an entropy estimator. The working mechanism of this estimator is straightforward and is detailed in \Cref{alg:compute_entropy}.

\begin{algorithm}[H]
\caption{\textsc{info-sedd-h}: Estimate $H(\fproc_0)$}
\label{alg:compute_entropy}
\begin{algorithmic}[1]
{\small
  \REQUIRE Initial sample $\fproc_0\sim \fp_0$, score network $\scorefn$
    \STATE $t\sim u(0,T)$
    \COMMENT{Sample time uniformly}
    \STATE $\fproc_t\sim \fp_t(\cdot | \fproc_0)$
    \COMMENT{Perturb data}
    \STATE $\hat{H}=0$
    \FOR{$i: \fproc^i_t=\absorb$}
    \FOR{$n\in [1:N]$}
    \STATE $\tilde{X}=[\fproc^1_t,\dots,\fproc^{i-1}_t,n,\fproc^{i+1}_t,\dots,\fproc^{M}_t]$
    \STATE $\hat{H}+=T\sigma(t)\left(K\left(\scorefn(\fproc_t)_{\tilde{X}}\right)+\ratio-\scorefn(\fproc_t)_{\tilde{X}}\log\left(\ratio\right)\right)$
    \ENDFOR
    \ENDFOR
    \RETURN $\hat{H}$
}
    \end{algorithmic}
\end{algorithm}


%% file: experiments.tex
In this section, we numerically validate the performance of \infosedd\ on both synthetic and real-world datasets. Specifically, we evaluate the mutual information and entropy estimators presented in \Cref{alg:infosedd} and \Cref{alg:compute_entropy}, respectively, through the following experiments: (i) benchmarking on high-dimensional distributions where ground truth values are known by construction, and (ii) assessing the accuracy of our method in a \textit{real-world} application by estimating the entropy of spin glass configurations in the Ising model \citep{onsager1944crystal}.

\subsection{Synthetic benchmark}\label{sec:synthetic}

\input{synthetic}

\paragraph{Experimental Setup} To evaluate the performance of our mutual information estimator, \infosedd, we design three sets of controlled synthetic experiments, each isolating a key aspect of data complexity. In the first experiment, we fix the mutual information value at $0.5$ and the length of the random vector at $2$, meaning each of the two random variables has one dimension. We then increase the support size using Cantor's mapping (\Cref{sec:isomorphisms}). The second experiment maintains a mutual information value of $0.5$ and binary support for each element of the random vector, but instead increases the length of the representation vector. Finally, in the third experiment, we generate distributions with varying mutual information values while keeping the support binary for each variable. The length of the random vectors is fixed at $10$, and mutual information values are linearly spaced from $0$ to $5$. We compare the results of our proposed methodology against MINE \citep{belghazi2018mine}, a variational neural estimator, MINDE \citep{franzese2023minde}, a generative neural estimator and KSG \citep{kraskov2004estimating}, a \textit{classical} statistical estimator. For a fair comparison, we use \glspl{MLP} based architectures for the neural estimators and scale parameters appropriately when needed (see \Cref{sec:training_details}). When an estimate is not available, we fill the entry in the tables with $-$. We report all results in nats.

\input{tables_and_figure}

\paragraph{Results and Analysis} The results from these experiments, shown in \Cref{tab:mi_big_support_table,tab:mi_big_vector_table,tab:mi_estimators}, demonstrate that \infosedd\ consistently outperforms competing methods. Notably, our model excels in settings where an inherent property of discrete systems—such as support size, representation length, or mutual information—introduces increased complexity.

Although our competitors perform relatively well in some benchmarks, they fail in at least one experiment. KSG performs well with large support size (\Cref{tab:mi_big_support_table}), but it fails when considering higher dimensions (\Cref{tab:mi_big_vector_table}). MINE and MINDE, on the contrary, excels with higher dimensions but struggle with large support size. MINDE also struggles more with higher mutual information values (\Cref{tab:mi_estimators}). Overall, these experiments motivate the usage of adequate neural estimators when dealing with discrete distributions. 

\subsection{Spin glasses experiments}
Entropy computation in Ising models enables insights on the thermodynamics properties of the system \citep{cincio2007entropy}, which can be used for scientific discovery in the domain where the Ising model is applied \citep{macy2024ising, schneidman2006weak, sherrington1975solvable}. 

\paragraph{Experimental setup} We consider a simplified Ising model applied to spin glasses \citep{sherrington1975solvable}. We do not include an external field, we set a unitary interaction strength for all the sites interactions, unitary Boltzmann's constant and a $20\times20$ square lattice. The entropy per site of this configuration can be computed in closed form \cite{onsager1944crystal}. We test our model by estimating the entropy per particle at linearly spaced temperatures from $1.0K$ to $4.0K$. We generate our dataset using the Metropolis-Hastings algorithm, with $10000$ samples for each temperature (see \Cref{subsec:app_ising}). We post-process the output of \infosedd\ by dividing the entropy estimate by $400$ to report the entropy per site.

\paragraph{Results and Analysis} Variational estimators cannot estimate large \glspl{KL} divergences reliably with limited samples sizes \citep{mcallester2020formal,song2019understanding}. In this scenario, instead, \infosedd\ accurately estimates large \glspl{KL} divergences (\Cref{fig:ising}), performing particularly well at low temperatures where we need to estimate large \glspl{KL} divergences.

%% file: synthetic.tex
Real-world data distributions grow rapidly in complexity. Modern language models \citep{radford2019language,lou2024discrete} process sequences of thousands of tokens from vocabularies of tens of thousands values. These correspond to the structured discrete variables introduced earlier, where tokens map to sequence length $M$, and vocabulary size corresponds to the number of states per subcomponent $N$. As mutual information quantifies dependencies, its value increases rapidly with $N^M$, making estimation challenging.

To evaluate the feasibility of mutual information estimation in high-dimensional discrete settings, we design synthetic experiments that highlight the limitations of estimators originally developed for continuous variables. We control three key factors: support size ($N$), representation dimension ($M$), and mutual information value. By varying one factor at a time, we systematically assess whether continuous-variable estimators fail while \textsc{info-sedd} scales effectively.

We generate joint distributions for random variables $X, Y$ with user-defined mutual information and support sizes $\support_X, \support_Y$. Using an evolutionary strategy, we encode the joint distribution in a vector $g_a \in \mathbb{R}^{|\support_X||\support_Y|}$ and transform it into a valid probability distribution via normalization and reshaping: $\frac{g_a-\text{min}(g_a)\ones+\epsilon\ones}{\ones(g_a-\text{min}(g_a)\ones+\epsilon\ones)}$ where $\epsilon$ ensures full support. The mutual information, computable in closed form, serves as the selection criterion in the evolutionary process. For large $|\support_X|$ and $|\support_Y|$, the evolutionary strategy struggles. Instead, we generate high mutual information distributions by concatenating independent distributions, leveraging the additive property of mutual information. Additionally, isomorphisms like Cantor’s pairing function, $\pi(x,y) = \frac{1}{2}(x+y)(x+y+1)+y$, enables support expansion without altering mutual information, aiding consistency across experiments. Full details are in \Cref{sec:isomorphisms}.

%% file: tables_and_figure.tex
\begin{table}[t]
\label{tab:mi_comparison}
\vspace{1em} 
\begin{center}
\begin{minipage}{0.45\textwidth}
\tiny
   
    \centering
    \caption{Different support dimension $|\support|$ and MI=0.5}
    \begin{tabular}{c|cccc}
        \multicolumn{1}{c}{\bf $|\chi|$} & \multicolumn{1}{c}{\bf MINDE} & \multicolumn{1}{c}{\bf MINE} & \multicolumn{1}{c}{\bf I-SEDD} & \multicolumn{1}{c}{\bf KSG} \\ 
        \hline \\
        2 & 0.51 & 0.50 & 0.50 & 0.49 \\
        4 & 0.04 & 0.46 & 0.53 & 0.49 \\
        16 & 0.02 & 0.00 & 0.51 & 0.49 \\
        64 & 0.00 & -12644 & 0.54 & 0.48 \\
        256 & 0.00 & 0.00 & 0.58 & 0.44 \\
        1024 & 0.00 & - & 0.61 & 0.24 \\
    \end{tabular}
    \label{tab:mi_big_support_table}
\end{minipage}
\begin{minipage}{0.45\textwidth}
\tiny

    \centering
    \caption{Different random vector lengths and MI=0.5}
    \begin{tabular}{c|cccc}
        \multicolumn{1}{c}{\bf Length} & \multicolumn{1}{c}{\bf MINDE} & \multicolumn{1}{c}{\bf MINE} & \multicolumn{1}{c}{\bf I-SEDD} & \multicolumn{1}{c}{\bf KSG} \\ 
        \hline \\
        2 & 0.55 & 0.50 & 0.50 & 0.51 \\
        8 & 0.39 & 0.51 & 0.55 & 0.51 \\
        32 & 0.27 & 0.73 & 0.52 & 0.07 \\
        128 & 0.35 & 0.43 & 0.48 & 0.00 \\
        512 & 0.38 & 1.22 & 1.60 & 0.00 \\
        2048 & 5.56 & 0.00 & 5.19 & 0.01 \\
    \end{tabular}
    \label{tab:mi_big_vector_table}
\end{minipage}


\begin{minipage}{0.45\textwidth}
\tiny

    \centering
    \caption{Different MI values}
    \begin{tabular}{c|cccc}
        \multicolumn{1}{c}{\bf MI} & \multicolumn{1}{c}{\bf MINDE} & \multicolumn{1}{c}{\bf MINE} & \multicolumn{1}{c}{\bf I-SEDD} & \multicolumn{1}{c}{\bf KSG} \\ 
        \hline \\
        0 & 0.40 & 0.02 & 0.00 & 0.01 \\
        1 & 0.23 & 1.26 & 1.01 & 0.49 \\
        2 & 1.44 & 2.29 & 2.03 & 1.14 \\
        3 & 2.17 & 3.08 & 3.02 & 2.15 \\
        4 & 3.51 & 3.84 & 4.01 & 3.39 \\
        5 & 17.92 & 4.93 & 5.07 & 4.72 \\
    \end{tabular}
    \label{tab:mi_estimators}
\end{minipage}
\hfill
\begin{minipage}{0.45\textwidth}
\includegraphics[scale=0.3]{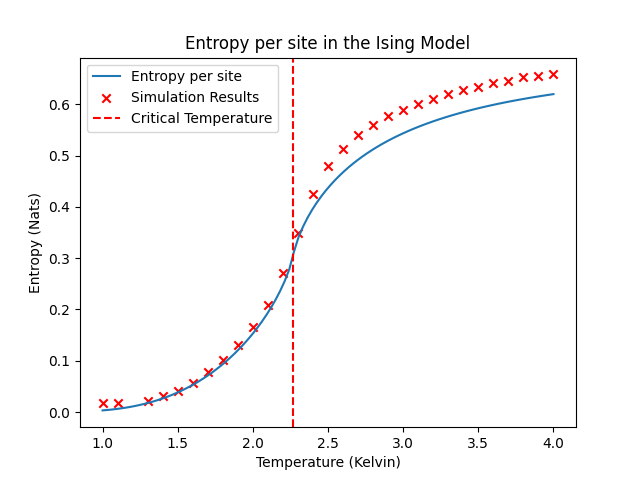}
\captionof{figure}{Entropy of the Ising model at different temperatures}\label{fig:ising}
\end{minipage}

\end{center}
\vspace{1em} 
\end{table}

%% file: conclusions.tex
Our work introduces \textsc{INFO-SEDD}, a novel method for estimating information-theoretic quantities in discrete data using \glspl{CTMC}. By leveraging a single parametric model, our approach offers computational and memory efficiency while maintaining accuracy in high-dimensional settings. Our experiments highlight \infosedd's robustness against existing neural estimators and demonstrate its effectiveness on both synthetic benchmarks and real-world applications, such as entropy estimation in the Ising model. These results underscore the importance of using specialized techniques for discrete distributions and open new avenues for scalable and accurate information-theoretic analysis in complex systems.

\section{Acknowledgments}
Alberto Foresti, Giulio Franzese and Pietro Michiardi were partially funded by project MUSECOM2 - AI-enabled MUltimodal SEmantic COMmunications and COMputing, in the Machine Learning-based Communication Systems, towards Wireless AI (WAI), Call 2022, ChistERA.



%% file: appendix.tex
\section{Appendix}

\subsection{Proof of \eqref{eq:infosedd}} We first break $\expected\left[\log\frac{\bp_T}{\bq_T}\bigg|\bproc_0\right]$ in $\expected\left[\log\bp_T\bigg|\bproc_0\right]-\expected\left[\log\bq_T\bigg|\bproc_0\right]$ and apply the Dynkin's formula separately. We start with $\expected\left[\log\bp_T\bigg|\bproc_0\right]$:

{\scriptsize\begin{align}
    \expected\left[\log\bp_T\bigg|\bproc_0\right] \nonumber&= \log\bp_0(\bproc_0)+\expected\left[\int_0^T\frac{\partial \log\bp_t}{\partial t}(\bproc_t,t)+\boperator[\log\bp_t](\bproc_t,t)dt\bigg|\bproc_0\right] \\
    & = \log\bp_0(\bproc_0)+\expected\left[\int_0^T\frac{\partial \log\bp_t}{\partial t}(\bproc_t,t)+\sum_{y\neq \bproc_t}\bratesp_t(y,\bproc_t)(\log\bp_t(y)-\log\bp_t(\bproc_t))dt\bigg|\bproc_0\right] \nonumber\\
    & = \log\bp_0(\bproc_0)+\expected\left[\int_0^T\frac{\partial \log\bp_t}{\partial t}(\bproc_t,t)+\sum_{y\neq \bproc_t}\bratesp_t(y,\bproc_t)\log\frac{\bp_t(y)}{\bp_t(\bproc_t)}dt\bigg|\bproc_0\right] \nonumber
\end{align}}

We now focus on the term $\frac{\partial \log\bp_t}{\partial t}$, which we can rewrite using the definition given in \eqref{eq:bctmc}:

\begin{align}
    \frac{\partial \log\bp_t}{\partial t} = \frac{\partial \bp_t}{\partial t}\bigg/ \bp_t = \frac{\bratesp_t\bp_t}{\bp_t}
\end{align}

Where the division between numerator and denominator in the last two terms denotes element-wise division. We focus now on simplifying the numerator, using the definition in \eqref{eq:ratesdef}. Recalling that

\begin{align}\label{eq:ratesdef}
    \bratesp_t(y,x) &= \left(\frac{\fp_{T-t}(y)}{\fp_{T-t}(x)} \fratesp_{T-t}(x,y)\right) (1-\delta(x,y))+\left(-\sum_{y\neq x}\bratesp_t(y,x)\right)\delta(x,x) \\ &= \left(\frac{\bp_{t}(y)}{\bp_{t}(x)} \fratesp_{T-t}(x,y)\right) (1-\delta(x,y))+\left(-\sum_{y\neq x}\bratesp_t(y,x)\right)\delta(x,x)
\end{align}

we compute the $x$-th element of $\bratesp_t\bp_t$:

\begin{align}
    [\bratesp_t\bp_t](x) &= \sum_y \bratesp_t(x,y)\bp_t(y) \\&= \sum_{y\neq x} \bratesp_t(x,y)\bp_t(y) + \bratesp_t(x,x)\bp_t(x)\nonumber\\
    &= \sum_{y\neq x} \fratesp_{T-t}(y,x)\frac{\bp_t(x)}{\bp_t(y)}\bp_t(y) - \sum_{y\neq x}\bratesp_t(y,x)\bp_t(x)\nonumber\\
    &= \sum_{y\neq x} \fratesp_{T-t}(y,x)\bp_t(x) - \sum_{y\neq x}\fratesp_{T-t}(x,y)\frac{\bp_t(y)}{\bp_t(x)}\bp_t(x)\nonumber\\
    &= \sum_{y\neq x} \fratesp_{T-t}(y,x)\bp_t(x) - \sum_{y\neq x}\fratesp_{T-t}(x,y)\bp_t(y)\nonumber
    \\
    &= \sum_{y\neq x} \fratesp_{T-t}(y,x)\bp_t(x) - \fratesp_{T-t}(x,y)\bp_t(y)\nonumber
\end{align}

Finally, if we divide by the denominator we obtain:

{\begin{align}
    \left[\frac{\bratesp_t\bp_t}{\bp_t}\right](x) = &\frac{\sum_{y\neq x} \fratesp_{T-t}(y,x)\bp_t(x) - \fratesp_{T-t}(x,y)\bp_t(y)}{\bp_t(x)} \\ = &  \sum_{y\neq x} \fratesp_{T-t}(y,x) - \fratesp_{T-t}(x,y)\frac{\bp_t(y)}{\bp_t(x)} \nonumber
\end{align}}

Moreover, if we notice that $\bratesp_t(y,\bproc_t)\log\frac{\bp_t(y)}{\bp_t(\bproc_t)}=\fratesp_{T-t}(\bproc_t,y)\frac{\bp_t(y)}{\bp_t(\bproc_t)}\log\frac{\bp_t(y)}{\bp_t(\bproc_t)}$, we can write the Dynkin's formula as:

{\begin{align}\label{eq:dynkinp}
    &\expected\left[\log\bp_T|\bproc_0\right] = \\ &\log\bp_0(\bproc_0)+ \nonumber\\ &\expected\left[\int_0^T\sum_{y\neq \bproc_t} \fratesp_{T-t}(y,\bproc_t) - \fratesp_{T-t}(\bproc_t,y)\frac{\bp_t(y)}{\bp_t(\bproc_t)}+\fratesp_{T-t}(\bproc_t,y)\frac{\bp_t(y)}{\bp_t(\bproc_t)}\log\frac{\bp_t(y)}{\bp_t(\bproc_t)}dt\bigg|\bproc_0\right] \nonumber
\end{align}}

Then, if we define $K(a)=a(\log a - 1)$ and group some terms we obtain:

\begin{align}
    \expected\left[\log\bp_T|\bproc_0\right] &= \\ \log\bp_0(\bproc_0)&+\expected\left[\int_0^T\sum_{y\neq \bproc_t} \fratesp_{T-t}(y,\bproc_t) + \fratesp_{T-t}(\bproc_t,y)K\left(\frac{\bp_t(y)}{\bp_t(\bproc_t)}\right)dt\bigg|\bproc_0\right] \nonumber
\end{align}

We now repeat similar calculations for $\expected\left[\log\bq_T\bigg|\bproc_0\right]$. Firstly, the term $\frac{\partial \log\bq_t}{\partial t}$ becomes:

\begin{equation}
    \left[\frac{\partial \log\bq_t}{\partial t}\right](x) = \sum_{y\neq x} \fratesp_{T-t}(y,x) - \fratesp_{T-t}(x,y)\frac{\bq_t(y)}{\bq_t(x)}
\end{equation}

Whereas the backward operator term $\boperator[\log\bq_t](\bproc_t,t)$ becomes:
\begin{equation}
    \bratesp_t(y,\bproc_t)\log\frac{\bq_t(y)}{\bq_t(\bproc_t)}=\fratesp_{T-t}(\bproc_t,y)\frac{\bp_t(y)}{\bp_t(\bproc_t)}\log\frac{\bq_t(y)}{\bq_t(\bproc_t)}
\end{equation}

Putting everything together gives:

{\begin{align}\label{eq:dynkinq}
    &\expected\left[\log\bq_T\bigg|\bproc_0\right] = \\ &\log\bq_0(\bproc_0)+ \nonumber\\&\expected\left[\int_0^T\sum_{y\neq \bproc_t} \fratesp_{T-t}(y,\bproc_t) - \fratesp_{T-t}(\bproc_t,y)\frac{\bq_t(y)}{\bq_t(\bproc_t)}+\fratesp_{T-t}(\bproc_t,y)\frac{\bp_t(y)}{\bp_t(\bproc_t)}\log\frac{\bq_t(y)}{\bq_t(\bproc_t)}dt\bigg|\bproc_0\right] \nonumber
\end{align}}

Finally, we can estimate $\expected\left[\log\frac{\bp_T}{\bq_T}\bigg|\bproc_0\right]$ by subtracting \eqref{eq:dynkinq} from \eqref{eq:dynkinp}:

{\scriptsize\begin{equation}
    \expected\left[\log\frac{\bp_T}{\bq_T}\bigg|\bproc_0\right] \approx \expected\left[\int_0^T\sum_{y\neq \bproc_t} \fratesp_{T-t}(\bproc_t,y)\left(K\left(\frac{\bp_t(y)}{\bp_t(\bproc_t)}\right)+\frac{\bq_t(y)}{\bq_t(\bproc_t)}-\frac{\bp_t(y)}{\bp_t(\bproc_t)}\log\frac{\bq_t(y)}{\bq_t(\bproc_t)}\right)dt\bigg|\bproc_0\right] \nonumber
\end{equation}}

By using the fact that $\expected\left[\expected\left[\log\frac{\bp_T}{\bq_T}\bigg|\bproc_0\right]\right]=\expected\left[\log\frac{\bp_T}{\bq_T}\right]$, $\bp_t=\fp_{T-t}$, $\bq_t=\fq_{T-t}$, $\bproc_t=\fproc_{T-t}$ and by setting $\tau=T-t$, we get:

\begin{align}
&\expected\left[\int_0^T\sum_{y\neq \fproc_\tau} \fratesp_{\tau}(\fproc_\tau,y)\left(K\left(\frac{\fp_\tau(y)}{\fp_\tau(\fproc_\tau)}\right)+\frac{\fq_\tau(y)}{\fq_\tau(\fproc_\tau)}-\frac{\fp_\tau(y)}{\fp_\tau(\fproc_\tau)}\log\frac{\fq_\tau(y)}{\fq_\tau(\fproc_\tau)}\right)d\tau\right]
\end{align}

recovering \Cref{eq:infosedd}.

\subsection{Proof $\frac{\fu_t(x)}{\fu_t(\absorb)}=\ratio$ for $x\neq\absorb$}\label{sec:absorb_ratio}

{\small\begin{align}
    \frac{\fu_t(x)}{\fu_t(\absorb)} = \frac{\sum_{x_0\in\support}\fu_t(x|x_0)\fu_0(x_0)}{\sum_{x_0\in\support}\fu_t(\absorb|x_0)\fu_0(x_0)}
    = \frac{\sum_{x_0\in\support}\delta(x,x_0)e^{-\cumnoise(t)}\frac{1}{N}}{\sum_{x_0\in\support}(1-e^{-\cumnoise(t)})\frac{1}{N}} \nonumber = \frac{e^{-\cumnoise(t)}}{N(1-e^{-\cumnoise(t)})} = \ratio
\end{align}}

\subsection{Proof of \Cref{eq:free_marginal}}\label{sec:free_marginal}

Consider $\fp_t(\fproc_t=\xbar,\fprocy_t=\absorb)$:

\begin{align}\label{eq:marg_score}
    &\fp_t(\fproc_t=\xbar,\fprocy_t=\absorb) = \sum_{x,y}\prob(\fproc_t=\xbar, \fprocy_t=\absorb, \fproc_0=x,\fprocy_0=y)\\ &= \sum_{x,y}\underset{\pjump}{\underbrace{\prob(\fprocy_s=\absorb\g\fproc_t=\xbar,\fproc_0=x,\fprocy_0=y)}}\prob(\fproc_t=\xbar\g\fproc_0=x,\fprocy_0=y)\prob(\fproc_0=x,\fprocy_0=y) \nonumber \\ &= \sum_{x,y}\pjump\prob(\fproc_t=\xbar\g\fproc_0=x)\prob(\fproc_0=x,\fprocy_0=y) \nonumber \\ &= \pjump\sum_x\prob(\fproc_t=\xbar\g\fproc_0=x)\underset{\prob(\fproc_0=x)}{\underbrace{\sum_y\prob(\fproc_0=x,\fprocy_0=y)}} \nonumber \\ & = \pjump\prob(\fproc_t=\xbar) \nonumber
\end{align}

\Cref{eq:marg_score} implies that $\frac{\fp_t(\fproc_t=x,\fprocy_t=\absorb)}{\fp_t(\fproc_t=\xbar,\fprocy_t=\absorb)} = \frac{\prob(\fproc_t=x)}{\prob(\fproc_t=\xbar)}$. This important property enables the estimation of mutual information without modifying the score network.

\section{Experimental details}

\subsection{Dataset generation}\label{sec:isomorphisms}
In our experiments, we exploit the additivity of mutual information with independent random variables to generate complex datasets. By appending discrete noise random variables $Z_\rvx$, $Z_\rvy$ to the original random variables $\fproc_0$, $\fprocy_0$, we have $I(\fproc_0,\fprocy_0) = I([\fproc_0, Z_\rvx],[\fprocy_0, Z_\rvy])$. Pairing functions are isomorphisms that map $\sN\times\sN$ to $\sN$. They allow preserving mutual information through the Markov Chain:

\begin{equation}
    [\fproc_0,Z_\rvx] \to \hat{X}_0 \to \hat{Y}_0 \to [\fprocy_0,Z_\rvy] \to \hat{Y}_0 \to \hat{X}_0 \to [\fproc_0,Z_\rvx]
\end{equation}

Where, by the \textit{data processing inequality}, we have $I([\fproc_0,Z_\rvx],[\fprocy_0,Z_\rvy]) \geq I(\hat{X}_0,\hat{Y}_0) \geq I([\fproc_0,Z_\rvx],[\fprocy_0,Z_\rvy]) \implies I([\fproc_0,Z_\rvx],[\fprocy_0,Z_\rvy]) = I(\hat{X}_0,\hat{Y}_0) \implies I(\hat{X}_0,\hat{Y}_0) = I(\fproc_0,\fprocy_0)$.

In \Cref{sec:synthetic} we keep the same support dimension for both random variables, increasing it using the following procedure:
\begin{enumerate}
    \item We sample two binomial random variables $Z_\rvx,Z_\rvy$ with parameters $(n,p)$, 
    We vary $n$ for increasing the complexity of the experiment and we keep $p$ fixed to $0.5$.
    \item We concatenate $Z_\rvx$ and $Z_\rvy$ respectively to $\fproc_0$ and $\fprocy_0$, to form higher support versions $\hat{X}_0,\hat{Y}_0$.
    \item We map the noisy $\hat{X}_0,\hat{Y}_0$ versions to univariate random variables by applying Cantor's mapping.
\end{enumerate}

\subsection{Model and training setup}\label{sec:training_details} For our method, we use a Multi Layer Perceptron (MLP) with skip connection, based on the architecture used in \cite{franzese2023minde}, reworking the initial layer to include absolute positional embeddings. For training, we match the methodology used by \cite{lou2024discrete}, using the absorb configuration. Similarly, we follow prior work to train other models in the synthetic benchmark. At inference time, we always take the last valid validation step estimate of each model to avoid not-a-number values in our tables.

\subsubsection{Synthetic benchmark} In this section, we describe how we scaled the architectures with increasing complexity in the synthetic benchmark.
We benchmark our competitors using architectures from prior work \citep{franzese2023minde, belghazi2018mine}. Depending on the task, some architectures are forced to increase the number of parameters. For example, \textsc{Info-SEDD} is forced to increase the number of parameters with the support size. In order to maintain a fair comparison between different models, we make the other architectures match the number of parameters of the architecture which is forced to include more parameters by either increasing the number of layers or by increasing the layer width.

\paragraph{Big support experiments} For supports smaller than $256$, we use architectures with a comparable number of parameters, in the order of $20k$. Instead, after this support dimension we increase the number of parameters, with an order of $70k$ parameters for support dimension $256$ and an order of $300K$ for support dimension $1024$.

\paragraph{Representation length experiments}For vectors of length shorter than $512$, we use architectures with a comparable number of parameters, remaining in the order of $20k$. Instead, after this length we increase the number of parameters, with an order of $50k$ parameters for length $512$ and an order of $150k$ for length $2048$.

\paragraph{High mutual information experiments} In this case we keep the number of parameters fixed on the order of $20k$, as fixed representation length and fixed support dimension allow MINE and MINDE to keep the number of parameters constant. 

\subsubsection{Ising model experiments}\label{subsec:app_ising} 

The Ising model is a system consisting of particles arranged in a lattice. In our experiments, we consider a $L\times L$ square. A particle $i$ of the lattice is associated with a discrete value $\sigma_{i}\in\{-1,+1\}$ called spin and each pair of particles $ij$ is characterized by an interaction strength $J_{ij}$. With no external fields, these quantities determine the energy $E(\sigma)$ of the configuration $\sigma$: 

\begin{equation}
E(\sigma)=\sum_{i,j}J_{ij}\sigma_i\sigma_j
\end{equation}

In turns, the energy of a configuration determines its likelihood. In particular, the configurations of the Ising model follow a probability distribution $\fp_0$ parametrised by the temperature $T$, the Boltzmann constant $\boltzmann$ and the interaction strengths:

\begin{equation}
    \fp_0(\sigma) = \frac{e^{-\beta E(\sigma)}}{Z(T)}
\end{equation}

Where $Z(T)=\sum_{i}e^{-\beta E(\sigma_i)}$ and $\beta=(\boltzmann T)^{-1}$. In order to generate our dataset from $\fp_0$, we follow the Metropolis algorithm \citep{bhanot1988metropolis}:

\begin{algorithm}[H]
\caption{Metropolis Algorithm for 2D Ising Spin Glass}
\begin{algorithmic}[1]
\STATE \textbf{Input:} Lattice size \( N \), interaction strengths \( J_{ij} \), temperature \( T \), number of iterations \( \text{iter\_max} \)
\STATE \textbf{Initialize:} Spin lattice \( \sigma \) with \( \sigma_{i,j} \in \{-1, +1\} \) randomly assigned
\FOR{iteration = 1 to \( \text{iter\_max} \)}
    \STATE Randomly select a lattice site \(i \)
    \STATE Compute the change in energy \( \Delta E \) if \( \sigma_{i,j} \) is flipped:
    \[
    \Delta E = 2 \, \sigma_i \sum_{j} J_{i,j} \, \sigma_{j}
    \]
    \STATE Generate a random number \( r \) uniformly distributed in \( [0, 1] \)
    \IF{ \( r < \exp\left(-\beta\Delta E\right) \) }
        \STATE Flip the spin: \( \sigma_i \leftarrow -\sigma_i \)
    \ENDIF
\ENDFOR
\STATE \textbf{Output:} Final spin configuration \( \sigma \)
\end{algorithmic}
\end{algorithm}

We compute the entropy of $\fp_0$ analytically, starting from the free energy $F$ per site of the lattice:

\begin{equation}\label{eq:free_energy}
    F(T) = -\boltzmann T\log\pf_T
\end{equation}
Where $\pf_T$ is the partition function, which depends on the interaction horizontal and vertical interaction strength. For simplicity, we consider the same interaction strength $J=1$ for all neighboring particles, while we set it to zero for non neighboring particles. Under these assumptions, we can calculate $\log\lambda$ with a double integral\citep{onsager1944crystal}:

{\footnotesize\begin{equation}
    \log\pf_T = \log 2 + \frac{1}{2\pi^2}\int_0^\pi\int_0^\pi\log(\cosh(2\beta J)\cosh(2\beta J)-\sinh(2\beta J)\cos(\theta_1)-\sinh(2\beta J)\cos(\theta_2))d\theta_1d\theta_2
\end{equation}}

For simplicity, we also set $\boltzmann=1$. From $F$, we can calculate the entropy $H$ using the thermodynamic relation $H=-\frac{\partial F}{\partial T}$. We compute the integral numerically using the \textit{SciPy} Python package \citep{2020SciPy-NMeth} and we approximate $H$ as $H\approx \frac{F(T+\Delta T)-F(T-\Delta T)}{2\Delta T}$, with $\Delta T = 10^{-4}$.

For what concerns the \textsc{Info-SEDD} architecture, we keep a single model configuration for all temperatures, both for the model, which contains around $90k$ parameters, and for the diffusion. To get the entropy per site, we divide the estimates of the model by the number of particles in the configurations (400).

%% file: main.bbl
\begin{thebibliography}{64}
\providecommand{\natexlab}[1]{#1}
\providecommand{\url}[1]{\texttt{#1}}
\expandafter\ifx\csname urlstyle\endcsname\relax
  \providecommand{\doi}[1]{doi: #1}\else
  \providecommand{\doi}{doi: \begingroup \urlstyle{rm}\Url}\fi

\bibitem[Alemi \& Fischer(2018)Alemi and Fischer]{alemi2019gilbo}
Alexander~A Alemi and Ian Fischer.
\newblock Gilbo: One metric to measure them all.
\newblock \emph{Advances in Neural Information Processing Systems}, 31, 2018.

\bibitem[Alemi et~al.(2016)Alemi, Fischer, Dillon, and Murphy]{alemi2019deep}
Alexander~A Alemi, Ian Fischer, Joshua~V Dillon, and Kevin Murphy.
\newblock Deep variational information bottleneck.
\newblock In \emph{International Conference on Learning Representations}, 2016.

\bibitem[Anderson(2012)]{anderson2012continuous}
William~J Anderson.
\newblock \emph{Continuous-time Markov chains: An applications-oriented approach}.
\newblock Springer Science \& Business Media, 2012.

\bibitem[Arimoto(1971)]{arimoto1971information}
Suguru Arimoto.
\newblock Information-theoretical considerations on estimation problems.
\newblock \emph{Information and control}, 19\penalty0 (3):\penalty0 181--194, 1971.

\bibitem[Austin et~al.(2021)Austin, Johnson, Ho, Tarlow, and Van Den~Berg]{austin2021structured}
Jacob Austin, Daniel~D Johnson, Jonathan Ho, Daniel Tarlow, and Rianne Van Den~Berg.
\newblock Structured denoising diffusion models in discrete state-spaces.
\newblock \emph{Advances in Neural Information Processing Systems}, 34:\penalty0 17981--17993, 2021.

\bibitem[Barber \& Agakov(2004)Barber and Agakov]{barber2004algorithm}
David Barber and Felix Agakov.
\newblock The im algorithm: a variational approach to information maximization.
\newblock \emph{Advances in neural information processing systems}, 16\penalty0 (320):\penalty0 201, 2004.

\bibitem[Belghazi et~al.(2018)Belghazi, Baratin, Rajeshwar, Ozair, Bengio, Courville, and Hjelm]{belghazi2018mine}
Mohamed~Ishmael Belghazi, Aristide Baratin, Sai Rajeshwar, Sherjil Ozair, Yoshua Bengio, Aaron Courville, and Devon Hjelm.
\newblock Mutual information neural estimation.
\newblock In \emph{Proceedings of the 35th International Conference on Machine Learning}, 2018.

\bibitem[Bell \& Sejnowski(1995)Bell and Sejnowski]{bell1995information}
Anthony~J Bell and Terrence~J Sejnowski.
\newblock An information-maximization approach to blind separation and blind deconvolution.
\newblock \emph{Neural computation}, 7\penalty0 (6):\penalty0 1129--1159, 1995.

\bibitem[Bhanot(1988)]{bhanot1988metropolis}
Gyan Bhanot.
\newblock The metropolis algorithm.
\newblock \emph{Reports on Progress in Physics}, 51\penalty0 (3):\penalty0 429, 1988.

\bibitem[Bounoua et~al.(2024)Bounoua, Franzese, and Michiardi]{bounouas}
Mustapha Bounoua, Giulio Franzese, and Pietro Michiardi.
\newblock S $omega$ i: Score-based o-information estimation.
\newblock In \emph{Forty-first International Conference on Machine Learning}, 2024.

\bibitem[Brekelmans et~al.(2022)Brekelmans, Huang, Ghassemi, Steeg, Grosse, and Makhzani]{brekelmans2023improving}
Rob Brekelmans, Sicong Huang, Marzyeh Ghassemi, Greg~Ver Steeg, Roger~Baker Grosse, and Alireza Makhzani.
\newblock Improving mutual information estimation with annealed and energy-based bounds.
\newblock In \emph{International Conference on Learning Representations}, 2022.

\bibitem[Butakov et~al.(2024)Butakov, Tolmachev, Malanchuk, Neopryatnaya, and Frolov]{butakov2024mutual}
Ivan Butakov, Alexander Tolmachev, Sofia Malanchuk, Anna Neopryatnaya, and Alexey Frolov.
\newblock Mutual information estimation via normalizing flows.
\newblock In \emph{The Thirty-eighth Annual Conference on Neural Information Processing Systems}, 2024.
\newblock URL \url{https://openreview.net/forum?id=JiQXsLvDls}.

\bibitem[Campbell et~al.(2022)Campbell, Benton, De~Bortoli, Rainforth, Deligiannidis, and Doucet]{campbell2022continuous}
Andrew Campbell, Joe Benton, Valentin De~Bortoli, Thomas Rainforth, George Deligiannidis, and Arnaud Doucet.
\newblock A continuous time framework for discrete denoising models.
\newblock \emph{Advances in Neural Information Processing Systems}, 35:\penalty0 28266--28279, 2022.

\bibitem[Chai et~al.(2009)Chai, Walther, Beck, and Fei-Fei]{chai2009exploring}
Barry Chai, Dirk Walther, Diane Beck, and Li~Fei-Fei.
\newblock Exploring functional connectivities of the human brain using multivariate information analysis.
\newblock \emph{Advances in neural information processing systems}, 22, 2009.

\bibitem[Chen et~al.(2016)Chen, Duan, Houthooft, Schulman, Sutskever, and Abbeel]{chen2016infogan}
Xi~Chen, Yan Duan, Rein Houthooft, John Schulman, Ilya Sutskever, and Pieter Abbeel.
\newblock Infogan: Interpretable representation learning by information maximizing generative adversarial nets.
\newblock \emph{Advances in neural information processing systems}, 29, 2016.

\bibitem[Cincio et~al.(2007)Cincio, Dziarmaga, Rams, and Zurek]{cincio2007entropy}
Lukasz Cincio, Jacek Dziarmaga, Marek~M Rams, and Wojciech~H Zurek.
\newblock Entropy of entanglement and correlations induced by a quench: Dynamics of a quantum phase transition in the quantum ising model.
\newblock \emph{Physical Review A—Atomic, Molecular, and Optical Physics}, 75\penalty0 (5):\penalty0 052321, 2007.

\bibitem[Darrin et~al.(2024)Darrin, Formont, Cheung, and Piantanida]{darrin2024textttcosmicmutualinformationtaskagnostic}
Maxime Darrin, Philippe Formont, Jackie Chi~Kit Cheung, and Pablo Piantanida.
\newblock $\texttt{COSMIC}$: Mutual information for task-agnostic summarization evaluation, 2024.
\newblock URL \url{https://arxiv.org/abs/2402.19457}.

\bibitem[Dubey et~al.(2024)Dubey, Jauhri, Pandey, Kadian, Al-Dahle, Letman, Mathur, Schelten, Yang, Fan, et~al.]{dubey2024llama}
Abhimanyu Dubey, Abhinav Jauhri, Abhinav Pandey, Abhishek Kadian, Ahmad Al-Dahle, Aiesha Letman, Akhil Mathur, Alan Schelten, Amy Yang, Angela Fan, et~al.
\newblock The llama 3 herd of models.
\newblock \emph{arXiv preprint arXiv:2407.21783}, 2024.

\bibitem[Eckford et~al.(2016)Eckford, Loparo, and Thomas]{Eckford_2016}
Andrew~W. Eckford, Kenneth~A. Loparo, and Peter~J. Thomas.
\newblock Finite-state channel models for signal transduction in neural systems.
\newblock In \emph{2016 IEEE International Conference on Acoustics, Speech and Signal Processing (ICASSP)}, pp.\  6300–6304. IEEE, March 2016.
\newblock \doi{10.1109/icassp.2016.7472889}.
\newblock URL \url{http://dx.doi.org/10.1109/ICASSP.2016.7472889}.

\bibitem[Ethayarajh et~al.(2022)Ethayarajh, Choi, and Swayamdipta]{ethayarajh2022understandingdatasetdifficultymathcalvusable}
Kawin Ethayarajh, Yejin Choi, and Swabha Swayamdipta.
\newblock Understanding dataset difficulty with $\mathcal{V}$-usable information, 2022.
\newblock URL \url{https://arxiv.org/abs/2110.08420}.

\bibitem[Federici et~al.(2023)Federici, Ruhe, and Forr{\'e}]{federici2023effectiveness}
Marco Federici, David Ruhe, and Patrick Forr{\'e}.
\newblock On the effectiveness of hybrid mutual information estimation.
\newblock \emph{arXiv preprint arXiv:2306.00608}, 2023.

\bibitem[Franzese et~al.(2023)Franzese, Bounoua, and Michiardi]{franzese2023minde}
Giulio Franzese, Mustapha Bounoua, and Pietro Michiardi.
\newblock Minde: Mutual information neural diffusion estimation.
\newblock \emph{arXiv preprint arXiv:2310.09031}, 2023.

\bibitem[Gao et~al.(2015)Gao, Ver~Steeg, and Galstyan]{gao2015efficient}
Shuyang Gao, Greg Ver~Steeg, and Aram Galstyan.
\newblock Efficient estimation of mutual information for strongly dependent variables.
\newblock In \emph{Artificial intelligence and statistics}, pp.\  277--286. PMLR, 2015.

\bibitem[Hanson(2007)]{hanson2007applied}
Floyd~B Hanson.
\newblock \emph{Applied stochastic processes and control for jump-diffusions: modeling, analysis and computation}.
\newblock SIAM, 2007.

\bibitem[Hjelm et~al.(2019)Hjelm, Fedorov, Lavoie-Marchildon, Grewal, Bachman, Trischler, and Bengio]{hjelm2019learning}
R~Devon Hjelm, Alex Fedorov, Samuel Lavoie-Marchildon, Karan Grewal, Phil Bachman, Adam Trischler, and Yoshua Bengio.
\newblock Learning deep representations by mutual information estimation and maximization.
\newblock In \emph{International Conference on Learning Representations}, 2019.

\bibitem[Ho et~al.(2020)Ho, Jain, and Abbeel]{ho2020}
Jonathan Ho, Ajay Jain, and Pieter Abbeel.
\newblock Denoising diffusion probabilistic models.
\newblock In H.~Larochelle, M.~Ranzato, R.~Hadsell, M.F. Balcan, and H.~Lin (eds.), \emph{Advances in Neural Information Processing Systems}, volume~33, pp.\  6840--6851. Curran Associates, Inc., 2020.

\bibitem[Holderrieth et~al.(2024)Holderrieth, Havasi, Yim, Shaul, Gat, Jaakkola, Karrer, Chen, and Lipman]{holderrieth2024generator}
Peter Holderrieth, Marton Havasi, Jason Yim, Neta Shaul, Itai Gat, Tommi Jaakkola, Brian Karrer, Ricky~TQ Chen, and Yaron Lipman.
\newblock Generator matching: Generative modeling with arbitrary markov processes.
\newblock \emph{arXiv preprint arXiv:2410.20587}, 2024.

\bibitem[Huang et~al.(2020)Huang, Makhzani, Cao, and Grosse]{huang2020evaluating}
Sicong Huang, Alireza Makhzani, Yanshuai Cao, and Roger Grosse.
\newblock Evaluating lossy compression rates of deep generative models.
\newblock In \emph{International Conference on Machine Learning}. PMLR, 2020.

\bibitem[Karbowski(2024)]{Karbowski_2024}
Jan Karbowski.
\newblock Information thermodynamics: From physics to neuroscience.
\newblock \emph{Entropy}, 26\penalty0 (9):\penalty0 779, September 2024.
\newblock ISSN 1099-4300.
\newblock \doi{10.3390/e26090779}.
\newblock URL \url{http://dx.doi.org/10.3390/e26090779}.

\bibitem[Kelly(1981)]{kelly1981reversibility}
Frank~P Kelly.
\newblock Reversibility and stochastic networks.
\newblock \emph{PF Kelly—New York: Willy}, 1981.

\bibitem[Kong et~al.(2022)Kong, Brekelmans, and Ver~Steeg]{kong2023information}
Xianghao Kong, Rob Brekelmans, and Greg Ver~Steeg.
\newblock Information-theoretic diffusion.
\newblock In \emph{International Conference on Learning Representations}, 2022.

\bibitem[Kraskov et~al.(2004)Kraskov, St{\"o}gbauer, and Grassberger]{kraskov2004estimating}
Alexander Kraskov, Harald St{\"o}gbauer, and Peter Grassberger.
\newblock Estimating mutual information.
\newblock \emph{Physical review E}, 69\penalty0 (6):\penalty0 066138, 2004.

\bibitem[Lee \& Rhee(2024)Lee and Rhee]{lee2024benchmarksuiteevaluatingneural}
Kyungeun Lee and Wonjong Rhee.
\newblock A benchmark suite for evaluating neural mutual information estimators on unstructured datasets, 2024.
\newblock URL \url{https://arxiv.org/abs/2410.10924}.

\bibitem[Letizia \& Tonello(2022)Letizia and Tonello]{letizia2022copula}
Nunzio~A Letizia and Andrea~M Tonello.
\newblock Copula density neural estimation.
\newblock \emph{arXiv preprint arXiv:2211.15353}, 2022.

\bibitem[Letizia et~al.(2023)Letizia, Novello, and Tonello]{letizia2023variational}
Nunzio~A Letizia, Nicola Novello, and Andrea~M Tonello.
\newblock Variational $ f $-divergence and derangements for discriminative mutual information estimation.
\newblock \emph{arXiv preprint arXiv:2305.20025}, 2023.

\bibitem[Lou et~al.(2024)Lou, Meng, and Ermon]{lou2024discrete}
Aaron Lou, Chenlin Meng, and Stefano Ermon.
\newblock Discrete diffusion modeling by estimating the ratios of the data distribution.
\newblock \emph{stat}, 1050:\penalty0 21, 2024.

\bibitem[MacKay(2003)]{mackay2003information}
David~JC MacKay.
\newblock \emph{Information theory, inference and learning algorithms}.
\newblock Cambridge university press, 2003.

\bibitem[Macy et~al.(2024)Macy, Szymanski, and Ho{\l}yst]{macy2024ising}
Michael~W Macy, Boleslaw~K Szymanski, and Janusz~A Ho{\l}yst.
\newblock The ising model celebrates a century of interdisciplinary contributions.
\newblock \emph{npj Complexity}, 1\penalty0 (1):\penalty0 10, 2024.

\bibitem[McAllester \& Stratos(2020)McAllester and Stratos]{mcallester2020formal}
David McAllester and Karl Stratos.
\newblock Formal limitations on the measurement of mutual information.
\newblock In \emph{International Conference on Artificial Intelligence and Statistics}, 2020.

\bibitem[Moon et~al.(1995)Moon, Rajagopalan, and Lall]{moon1995estimation}
Young-Il Moon, Balaji Rajagopalan, and Upmanu Lall.
\newblock Estimation of mutual information using kernel density estimators.
\newblock \emph{Physical Review E}, 52\penalty0 (3):\penalty0 2318, 1995.

\bibitem[Newcomb \& Sayood(2021)Newcomb and Sayood]{newcomb2021use}
Garin Newcomb and Khalid Sayood.
\newblock Use of average mutual information and derived measures to find coding regions.
\newblock \emph{Entropy}, 23\penalty0 (10):\penalty0 1324, 2021.

\bibitem[Nguyen et~al.(2007)Nguyen, Wainwright, and Jordan]{nguyen2007neurips}
XuanLong Nguyen, Martin~J Wainwright, and Michael Jordan.
\newblock Estimating divergence functionals and the likelihood ratio by penalized convex risk minimization.
\newblock In \emph{Advances in Neural Information Processing Systems}, 2007.

\bibitem[Nowozin et~al.(2016)Nowozin, Cseke, and Tomioka]{nowozin2016neurips}
Sebastian Nowozin, Botond Cseke, and Ryota Tomioka.
\newblock F-gan: Training generative neural samplers using variational divergence minimization.
\newblock In \emph{Proceedings of the 30th International Conference on Neural Information Processing Systems}, NIPS'16, pp.\  271–279, Red Hook, NY, USA, 2016. Curran Associates Inc.
\newblock ISBN 9781510838819.

\bibitem[Onsager(1944)]{onsager1944crystal}
Lars Onsager.
\newblock Crystal statistics. i. a two-dimensional model with an order-disorder transition.
\newblock \emph{Physical Review}, 65\penalty0 (3-4):\penalty0 117, 1944.

\bibitem[Oord et~al.(2018)Oord, Li, and Vinyals]{oord2019representation}
Aaron van~den Oord, Yazhe Li, and Oriol Vinyals.
\newblock Representation learning with contrastive predictive coding.
\newblock \emph{Advances in neural information processing systems}, 2018.

\bibitem[Papamakarios et~al.(2017)Papamakarios, Pavlakou, and Murray]{papamakarios2017masked}
George Papamakarios, Theo Pavlakou, and Iain Murray.
\newblock Masked autoregressive flow for density estimation.
\newblock \emph{Advances in neural information processing systems}, 30, 2017.

\bibitem[Pinchas et~al.(2024)Pinchas, Ben-Gal, and Painsky]{pinchas2024comparative}
Assaf Pinchas, Irad Ben-Gal, and Amichai Painsky.
\newblock A comparative analysis of discrete entropy estimators for large-alphabet problems.
\newblock \emph{Entropy}, 26\penalty0 (5):\penalty0 369, 2024.

\bibitem[Pizer et~al.(1987)Pizer, Amburn, Austin, Cromartie, Geselowitz, Greer, ter Haar~Romeny, Zimmerman, and Zuiderveld]{pizer1987adaptive}
Stephen~M Pizer, E~Philip Amburn, John~D Austin, Robert Cromartie, Ari Geselowitz, Trey Greer, Bart ter Haar~Romeny, John~B Zimmerman, and Karel Zuiderveld.
\newblock Adaptive histogram equalization and its variations.
\newblock \emph{Computer vision, graphics, and image processing}, 39\penalty0 (3):\penalty0 355--368, 1987.

\bibitem[Poole et~al.(2019)Poole, Ozair, Van Den~Oord, Alemi, and Tucker]{poole2019variational}
Ben Poole, Sherjil Ozair, Aaron Van Den~Oord, Alex Alemi, and George Tucker.
\newblock On variational bounds of mutual information.
\newblock In \emph{International Conference on Machine Learning}, 2019.

\bibitem[Radford et~al.(2019)Radford, Wu, Child, Luan, Amodei, Sutskever, et~al.]{radford2019language}
Alec Radford, Jeffrey Wu, Rewon Child, David Luan, Dario Amodei, Ilya Sutskever, et~al.
\newblock Language models are unsupervised multitask learners.
\newblock \emph{OpenAI blog}, 1\penalty0 (8):\penalty0 9, 2019.

\bibitem[Ren et~al.(2024)Ren, Chen, Rotskoff, and Ying]{ren2024discrete}
Yinuo Ren, Haoxuan Chen, Grant~M Rotskoff, and Lexing Ying.
\newblock How discrete and continuous diffusion meet: Comprehensive analysis of discrete diffusion models via a stochastic integral framework.
\newblock \emph{arXiv preprint arXiv:2410.03601}, 2024.

\bibitem[Rhodes et~al.(2020)Rhodes, Xu, and Gutmann]{rhodes2020telescoping}
Benjamin Rhodes, Kai Xu, and Michael~U Gutmann.
\newblock Telescoping density-ratio estimation.
\newblock \emph{Advances in neural information processing systems}, 2020.

\bibitem[Schneidman et~al.(2006)Schneidman, Berry, Segev, and Bialek]{schneidman2006weak}
Elad Schneidman, Michael~J Berry, Ronen Segev, and William Bialek.
\newblock Weak pairwise correlations imply strongly correlated network states in a neural population.
\newblock \emph{Nature}, 440\penalty0 (7087):\penalty0 1007--1012, 2006.

\bibitem[Shannon(1948)]{shannon1948mathematical}
Claude~Elwood Shannon.
\newblock A mathematical theory of communication.
\newblock \emph{The Bell system technical journal}, 27\penalty0 (3):\penalty0 379--423, 1948.

\bibitem[Sherrington \& Kirkpatrick(1975)Sherrington and Kirkpatrick]{sherrington1975solvable}
David Sherrington and Scott Kirkpatrick.
\newblock Solvable model of a spin-glass.
\newblock \emph{Physical review letters}, 35\penalty0 (26):\penalty0 1792, 1975.

\bibitem[Song \& Ermon(2019{\natexlab{a}})Song and Ermon]{song2019understanding}
Jiaming Song and Stefano Ermon.
\newblock Understanding the limitations of variational mutual information estimators.
\newblock In \emph{International Conference on Learning Representations}, 2019{\natexlab{a}}.

\bibitem[Song \& Ermon(2019{\natexlab{b}})Song and Ermon]{song2019}
Yang Song and Stefano Ermon.
\newblock Generative modeling by estimating gradients of the data distribution.
\newblock In H.~Wallach, H.~Larochelle, A.~Beygelzimer, F.~d'~Alch\'{e}-Buc, E.~Fox, and R.~Garnett (eds.), \emph{Advances in Neural Information Processing Systems}, volume~32. Curran Associates, Inc., 2019{\natexlab{b}}.

\bibitem[Song et~al.(2021)Song, Sohl-Dickstein, Kingma, Kumar, Ermon, and Poole]{song2021a}
Yang Song, Jascha Sohl-Dickstein, Diederik~P Kingma, Abhishek Kumar, Stefano Ermon, and Ben Poole.
\newblock Score-based generative modeling through stochastic differential equations.
\newblock In \emph{International Conference on Learning Representations}, 2021.

\bibitem[Stratos(2019)]{stratos2018mutual}
Karl Stratos.
\newblock Mutual information maximization for simple and accurate part-of-speech induction.
\newblock In \emph{Proceedings of the 2019 Conference of the North American Chapter of the Association for Computational Linguistics: Human Language Technologies, Volume 1 (Long and Short Papers)}, 2019.

\bibitem[Sun et~al.(2023)Sun, Yu, Dai, Schuurmans, and Dai]{sun2023scorebasedcontinuoustimediscretediffusion}
Haoran Sun, Lijun Yu, Bo~Dai, Dale Schuurmans, and Hanjun Dai.
\newblock Score-based continuous-time discrete diffusion models, 2023.
\newblock URL \url{https://arxiv.org/abs/2211.16750}.

\bibitem[Virtanen et~al.(2020)Virtanen, Gommers, Oliphant, Haberland, Reddy, Cournapeau, Burovski, Peterson, Weckesser, Bright, {van der Walt}, Brett, Wilson, Millman, Mayorov, Nelson, Jones, Kern, Larson, Carey, Polat, Feng, Moore, {VanderPlas}, Laxalde, Perktold, Cimrman, Henriksen, Quintero, Harris, Archibald, Ribeiro, Pedregosa, {van Mulbregt}, and {SciPy 1.0 Contributors}]{2020SciPy-NMeth}
Pauli Virtanen, Ralf Gommers, Travis~E. Oliphant, Matt Haberland, Tyler Reddy, David Cournapeau, Evgeni Burovski, Pearu Peterson, Warren Weckesser, Jonathan Bright, St{\'e}fan~J. {van der Walt}, Matthew Brett, Joshua Wilson, K.~Jarrod Millman, Nikolay Mayorov, Andrew R.~J. Nelson, Eric Jones, Robert Kern, Eric Larson, C~J Carey, {\.I}lhan Polat, Yu~Feng, Eric~W. Moore, Jake {VanderPlas}, Denis Laxalde, Josef Perktold, Robert Cimrman, Ian Henriksen, E.~A. Quintero, Charles~R. Harris, Anne~M. Archibald, Ant{\^o}nio~H. Ribeiro, Fabian Pedregosa, Paul {van Mulbregt}, and {SciPy 1.0 Contributors}.
\newblock {{SciPy} 1.0: Fundamental Algorithms for Scientific Computing in Python}.
\newblock \emph{Nature Methods}, 17:\penalty0 261--272, 2020.
\newblock \doi{10.1038/s41592-019-0686-2}.

\bibitem[Wunder et~al.(2021)Wunder, Gro{\ss}, Fritschek, and Schaefer]{wunder2021reverse}
Gerhard Wunder, Benedikt Gro{\ss}, Rick Fritschek, and Rafael~F Schaefer.
\newblock A reverse jensen inequality result with application to mutual information estimation.
\newblock In \emph{2021 IEEE Information Theory Workshop (ITW)}, 2021.

\bibitem[Xia et~al.()Xia, Chen, Zhou, Shan, Du, Gao, Tan, Hu, Zheng, and Li]{xiaadanovo}
Jun Xia, Shaorong Chen, Jingbo Zhou, Xiaojun Shan, Wenjie Du, Zhangyang Gao, Cheng Tan, Bozhen Hu, Jiangbin Zheng, and Stan~Z Li.
\newblock Adanovo: Towards robust$\backslash$emph $\{$De Novo$\}$ peptide sequencing in proteomics against data biases.
\newblock In \emph{The Thirty-eighth Annual Conference on Neural Information Processing Systems}.

\bibitem[Zhao et~al.(2018)Zhao, Song, and Ermon]{zhao2018information}
Shengjia Zhao, Jiaming Song, and Stefano Ermon.
\newblock A lagrangian perspective on latent variable generative models.
\newblock In \emph{Proc. 34th Conference on Uncertainty in Artificial Intelligence}, 2018.

\end{thebibliography}
